\documentclass{article}

\usepackage{microtype}
\usepackage{graphicx}
\usepackage{subfigure}
\usepackage{booktabs} 

\usepackage{hyperref}

\usepackage[accepted]{icml2025}

\usepackage{amsmath}
\usepackage{amssymb}
\usepackage{mathtools}
\usepackage{amsthm}
\usepackage{multirow}
\usepackage{xcolor}
\usepackage{booktabs}

\usepackage{graphicx}
\usepackage{hhline}

\usepackage[capitalize,noabbrev]{cleveref}

\theoremstyle{plain}

\theoremstyle{definition}

\theoremstyle{remark}

\usepackage[textsize=tiny]{todonotes}

\icmltitlerunning{Healing Powers of BERT: How Task-Specific Fine-Tuning Recovers Corrupted Language Models}

\begin{document}

\twocolumn[
\icmltitle{Healing Powers of BERT: 
\\
How Task-Specific Fine-Tuning Recovers Corrupted Language Models}



\icmlsetsymbol{equal}{*}

\begin{icmlauthorlist}
\icmlauthor{Shijie Han}{equal,yyy}
\icmlauthor{Zhenyu Zhang}{equal,comp}
\icmlauthor{Andrei Arsene Simion}{yyy}
\end{icmlauthorlist}

\icmlaffiliation{yyy}{Columbia University}
\icmlaffiliation{comp}{Zhejiang University}

\icmlcorrespondingauthor{Shijie Han}{sh4460@columbia.edu}

\icmlkeywords{Robustness, Hierarchical \& Concept explanations, Probing}

\vskip 0.3in
]



\printAffiliationsAndNotice{\icmlEqualContribution} 

\begin{abstract}
Language models like BERT excel at sentence classification tasks due to extensive pre-training on general data, but their robustness to parameter corruption is unexplored. To understand this better, we look at what happens if a language model is "broken", in the sense that some of its parameters are corrupted and then recovered by fine-tuning. Strategically corrupting BERT variants at different levels, we find corrupted models struggle to fully recover their original performance, with higher corruption causing more severe degradation. Notably, bottom-layer corruption affecting fundamental linguistic features is more detrimental than top-layer corruption. Our insights contribute to understanding language model robustness and adaptability under adverse conditions, informing strategies for developing resilient NLP systems against parameter perturbations.
\end{abstract}
\section{Introduction}
Pre-trained language models like BERT \cite{kenton2019bert} have propelled tremendous progress in sentence classification, an important natural language processing (NLP) task. The powerful representational capabilities of BERT enable more accurate and robust classification across diverse domains \cite{yin2019benchmarking}. While BERT demonstrates exceptional ability in capturing intricate linguistic nuances, enabling it to analyze entire sentences and discern their meanings and relationships \cite{reimers2019sentence}, the inner workings of BERT are still not fully understood.

Several studies have investigated the robustness of BERT to various perturbations \cite{dong2021should, hauser2023bert}, but the effects of parameter corruption on fine-tuning performance for specific tasks remain unexplored. In this paper, we explore the impact of controlled parameter corruption on BERT's ability to perform sentence classification tasks. We hypothesize that corrupting a pre-trained BERT model will hinder its fine-tuned performance on sentence classification tasks, and then we explore how BERT recovers its original performance during task-specific fine-tuning.

To build our understanding, we strategically corrupt a pre-trained BERT model and then fine-tune it on a popular set of sentence classification datasets. We compare the fine-tuned performance of the corrupted model with a non-corrupted BERT model fine-tuned on the same task. Given this setup, we hope that this will allow us to quantify the classification performance degradation caused by corruption and assess BERT's ability to mitigate this degradation through fine-tuning.

This work sheds light on several crucial aspects of BERT's inner workings:
\begin{itemize}
    \item \textbf{Evaluation of language model robustness: }We assess how parameter corruption affects the performance of BERT models across various sentence classification tasks. We demonstrate that corrupted models cannot fully recover their original performance through fine-tuning, with higher corruption levels leading to more degradation.
    \item \textbf{Identification of critical architecture: }We reveal that corrupting bottom layers of BERT models has a more detrimental effect on performance compared to top layers. This finding highlights the importance of lower layers in capturing fundamental linguistic features.
    \item \textbf{Insights into Generalizability: }By analyzing the impact of corruption, we gain insight into BERT's ability to generalize from pre-trained knowledge to new tasks despite corruption.
\end{itemize}

Our findings contribute to a deeper understanding of BERT's robustness and the corrective abilities of fine-tuning. This knowledge can inform strategies for improving model robustness and developing more resilient NLP systems.
\section{Related Work}\label{sec:bg}
BERT (Bidirectional Encoder Representations from Transformers) revolutionized NLP with its pre-training and fine-tuning methods \cite{kenton2019bert}. Unlike GPT-style unidirectional models, BERT uses a bidirectional transformer architecture, understanding word context from both directions. This bidirectional approach and its rich token representations enable BERT to excel in various natural language understanding tasks.

\paragraph{Pre-training of BERT.}
BERT pre-training involves two objectives: Masked Language Modeling (MLM) and Next Sentence Prediction (NSP). In MLM, some input tokens are randomly masked, and the model learns to predict the masked tokens using the surrounding context. This allows BERT to capture nuanced language patterns and dependencies by considering context from both sides of a token \cite{radford2018improving}. The NSP task further enhances BERT by training it to predict if a given sentence B logically follows sentence A \cite{liu2019roberta, sanh2019distilbert}. NSP helps model understand relationships between sentences, which is important for tasks involving sentence pairs \cite{peters2018deep}.

\paragraph{Fine-tuning for Text Classification.}
Once pre-trained, BERT can be fine-tuned for specific downstream tasks with minimal architectural modifications. For text classification, a simple linear layer is added on top of the pre-trained model, using the output of the special [CLS] token as the aggregate sequence representation \cite{kenton2019bert, sun2019fine}. During fine-tuning, the entire model is trained on labeled data for the target task. BERT's rich pre-trained knowledge provides a strong foundation that can accurately classify text based on nuanced linguistic features when fine-tuned on labeled datasets \cite{yin2019benchmarking}. The [CLS] token output is passed through an additional classification layer to produce the final output, making BERT highly versatile for various text classification applications like sentiment analysis and topic categorization \cite{wang2018glue, hoang2019aspect, xu2019bert, george2023integrated}.

\paragraph{Robustness of internal structure.}
Despite the impressive performance of the BERT model, its inherent complex architecture also makes it vulnerable to various forms of corruption, such as adversarial attacks and weight perturbations, which can severely impact its performance \cite{ebrahimi2018hotflip}. While some research has attempted targeted interventions to influence model performance based on an understanding of the Transformer architecture, our understanding of how its internal components react to corruption remains limited \cite{meng2022locating, dai2022knowledge, geva2021transformer, de2021editing}. This highlights the need for a deeper understanding of the model, as enhancing our comprehension of BERT's robustness is crucial for model interpretability and reliable deployment \cite{feder2021causalm}. To bridge this gap, we conduct an in-depth investigation to reveal the extent to which BERT can recover through fine-tuning after being corrupted. 
\section{Corruption Method}\label{sec:corruption}
The real-world corruption of BERT is likely to be random and unpredictable, rather than following the same distribution as the pre-trained parameters. Therefore, we simulate corruption in BERT by randomly initializing their internal parameters. Specifically, for corrupting the parameters of the \textit{LayerNorm} layers, we standardize all weight parameters by setting them to a constant value of 1.0. For the weights of \textit{non-LayerNorm} layers, we re-initialize them using the Kaiming uniform initialization method \cite{he2015delving}, which draws samples from a uniform distribution with specific bounds determined by the input size of the layer. Additionally, all bias parameters are directly initialized to zero. 

By selectively applying this initialization corruption to designated layers, we can assess the effects of corruption across different layer positions within the BERT model. Subsequently, we fine-tune and evaluate the modified language model across various text classification benchmarks. 

\begin{table}[t]
  \centering
  \caption{Classification datasets details}
  \resizebox{0.47\textwidth}{!}{
  \begin{tabular}{c|cccccc}
    \toprule
    \textbf{Dateset} & \textbf{Categories} & \textbf{Train} & \textbf{Test} & \textbf{Balanced} \\
    \midrule
    SST-2 & 2 & 56k & 12k & Yes\\
    IMDB & 2 & 25k & 5k & Yes\\
    AG News & 4 & 48k & 6k & Yes\\
    Emotion & 6 & 14k & 4k & No\\
    DBPedia & 14 & 50k & 12k & Yes\\
    TFNT & 20 & 17k & 4k & No\\
    \bottomrule
  \end{tabular}
  }
  
  \label{tab:datasets}
\end{table}
\begin{table*}[!t]
\centering
\caption{Fine-tuned BERT performance under different corruption settings across various datasets. Bold values represent higher values between bottom and top. Red values indicate outliers. Reported F1 score in percentage (\%)}
\resizebox{\linewidth}{!}
{
\begin{tabular}{cc|cc|cc|cc|cc|cc|cc}
\toprule
\multirow{2}{*}{Model} & \multirow{2}{*}{Corruption} & \multicolumn{2}{c}{SST-2} & \multicolumn{2}{c}{IMDB} & \multicolumn{2}{c}{AG News} & \multicolumn{2}{c}{Emotion} & \multicolumn{2}{c}{DBPedia} & \multicolumn{2}{c}{TFNT} \\ 

 &  & Bottom & \multicolumn{1}{c}{Top} & Bottom & \multicolumn{1}{c}{Top} & Bottom & \multicolumn{1}{c}{Top} & Bottom & \multicolumn{1}{c}{Top} & Bottom & \multicolumn{1}{c}{Top} & Bottom & \multicolumn{1}{c}{Top} \\ 
 
\midrule
\multirow{5}{*}{Base} & 0\% & \multicolumn{2}{c|}{92.11} & \multicolumn{2}{c|}{82.41} & \multicolumn{2}{c|}{94.15} & \multicolumn{2}{c|}{93.04} & \multicolumn{2}{c|}{98.78} & \multicolumn{2}{c}{89.42} \\ 

 & 25\% (1-3) & 90.16 & \textbf{91.72} & 81.79 & \textbf{82.34} & 93.41 & \textbf{94.06} & 92.87 & \textbf{93.03} & \textbf{98.72} & 98.57 & 86.68 & \textbf{89.16} \\
 & 50\% (1-6) & 83.25 & \textbf{87.92} & 76.91 & \textbf{80.92} & 91.82 & \textbf{93.59} & 92.11 & \textbf{92.90} & 97.40 & \textbf{98.37} & 84.42 & \textbf{88.08} \\
 & 75\% (1-9) & 82.18 & \textbf{84.63} & 76.26 & \textbf{79.30} & 92.25 & \textbf{93.03} & 90.31 & \textbf{92.76} & 97.17 & \textbf{98.22} & 84.29 & \textbf{86.28} \\
 
 & 100\% (1-12) & \multicolumn{2}{c|}{75.17} & \multicolumn{2}{c|}{76.12} & \multicolumn{2}{c|}{90.36} & \multicolumn{2}{c|}{88.09} & \multicolumn{2}{c|}{94.08} & \multicolumn{2}{c}{75.98} \\[0.5mm]
 
\hhline{~~------------}
\rule{0pt}{4.3mm}

\multirow{5}{*}{Large} & 0\% &  \multicolumn{2}{c|}{93.22} & \multicolumn{2}{c|}{84.52} & \multicolumn{2}{c|}{94.18} & \multicolumn{2}{c|}{92.39} & \multicolumn{2}{c|}{98.71} & \multicolumn{2}{c}{90.07} \\ 

 & 25\% (1-6) & 90.25 & \textbf{91.78} & 82.85 & \textbf{83.32} & 93.52 & \textbf{93.84} & \textcolor{red}{16.98} & \textbf{93.38} & 97.62 & \textbf{98.65} & 86.99 & \textbf{89.93} \\
 & 50\% (1-12) & 81.83 & \textbf{86.47} & \textcolor{red}{33.97} & \textbf{79.48} & 92.23 & \textbf{93.70} & \textcolor{red}{16.98} & \textbf{93.24} & 96.82 & \textbf{98.51} & 81.65 & \textbf{89.25} \\
 & 75\% (1-18) & 80.84 & \textbf{84.86} & 76.60 & \textbf{79.10} & 91.84 & \textbf{93.38} & 90.02 & \textbf{92.18} & 96.65 & \textbf{98.49} & 83.27 & \textbf{87.39} \\
 
 & 100\% (1-24) & \multicolumn{2}{c|}{78.93} & \multicolumn{2}{c|}{76.38} & \multicolumn{2}{c|}{90.66} & \multicolumn{2}{c|}{87.80} & \multicolumn{2}{c|}{93.02} & \multicolumn{2}{c}{77.62} \\[0.5mm]

\hhline{~~------------}
\rule{0pt}{4.3mm}
 
\multirow{3}{*}{Distil} & 0\% & \multicolumn{2}{c|}{89.72} & \multicolumn{2}{c|}{81.74} & \multicolumn{2}{c|}{93.95} & \multicolumn{2}{c|}{93.11} & \multicolumn{2}{c|}{98.72} & \multicolumn{2}{c}{89.10} \\ 

 & 50\% (1-3) & 82.63 & \textbf{83.91} & 75.70 & \textbf{79.00} & 92.38 & \textbf{93.48} & 92.26 & \textbf{93.04} & 97.91 & \textbf{98.50} & 84.55 & \textbf{86.89} \\
 
 & 100\% (1-6) & \multicolumn{2}{c|}{77.50} & \multicolumn{2}{c|}{75.67} & \multicolumn{2}{c|}{90.57} & \multicolumn{2}{c|}{88.30} & \multicolumn{2}{c|}{94.32} & \multicolumn{2}{c}{77.66} \\
 
 \bottomrule
\end{tabular}

}

\label{tab:corruption_cls}
\end{table*}

\section{Experiment}
\subsection{Experimental Settings}
\paragraph{Language Models and Datasets.}
To assess the robustness of language models for parameter corruption during task-specific fine-tuning, we choose three BERT models with different scales: BERT-Base with 110M parameters \cite{kenton2019bert}, BERT-Large with 340M parameters \cite{kenton2019bert}, and DistilBERT with 66M parameters \cite{sanh2019distilbert}. We corrupt these models using the method outlined in Section \ref{sec:corruption} and then fine-tune and evaluate them on six classification datasets spanning sentiment analysis and topic categorization: SST-2 \cite{socher-etal-2013-recursive}, IMBD \cite{maas-EtAl:2011:ACL-HLT2011}, AG News \cite{NIPS2015_250cf8b5}, Emotion \cite{saravia-etal-2018-carer}, DBPedia \cite{NIPS2015_250cf8b5} and Twitter Financial News Topic (TFNT) \cite{magic2022twitter}. To control our experiments, we balance the learning difficulty on a dataset with the available computational resources by truncating portions of the dataset and redefining the training and testing splits, as detailed in Table \ref{tab:datasets}.

\paragraph{Implementation Details.}
Our implementation utilizes the following software and hardware configurations: \textit{PyTorch} \cite{paszke2019pytorch} version 1.13.1; \textit{Transformers} \cite{wolf-etal-2020-transformers} library version 4.41.1; \textit{CUDA} version 12.4; \textit{GPU}: one NVIDIA RTX4090D-24G; We employ different learning rates of 1e-5, 2e-5, and 5e-5 and choose the best, training over 10 epochs for each dataset with Adam \cite{kingma2015adam} as our optimizer.

\subsection{Main Results}\label{sec:main}
To explore the impact of different positions and degrees of corruption on BERT recovery via fine-tuning, in addition to testing the original and fully corrupted models, we corrupt them by 25\%, 50\%, and 75\% from the bottom and top respectively (the DistilBERT only performs 50\% damage). To assess our experiments, we use the weighted F1 score, and in each case, we test every epoch and choose the best indicator. All results are shown in Table \ref{tab:corruption_cls}. Based on these experimental results, we can make the following analysis:

\textbf{1. Generally, a corrupted BERT model is unable to fully recover its original performance via fine-tuning. }Across all three BERT models and six datasets, the fine-tuned performance on corrupted models does not match the performance of the undamaged models. This trend persists across different corruption levels and datasets, suggesting that the fine-tuning process has limitations in compensating for parameter corruption.

\textbf{2. As the degree of corruption increases, the fine-tuned performance decreases. }When the models are undamaged (0\% corruption), they achieve the highest performance. As the corruption degree increases from 25\% to 50\%, 75\%, and 100\%, the F1 scores gradually decline, indicating a decreased ability to recover from the parameter corruption through fine-tuning. For instance, with the BERT-Base model on the SST-2 dataset, the F1 score drops from 92.11 (undamaged) to 90.16 (25\% bottom), 83.25 (50\% bottom), 82.18 (75\% bottom), and 75.17 (100\% corruption). Similar patterns can be observed for the other models and datasets. This clearly shows that corruption can cause irreparable damage to the model's performance and there is a direct observable pattern to this degradation.

\textbf{3. When the degree of corruption remains unchanged, the impact of bottom-layer corruption is greater than top-layer corruption. }Specifically, comparing performance between the bottom and top corruption at the same level, it is evident that bottom-layer corruption leads to a more substantial decrease in F1 score compared to the top layer. For example, with the BERT-Base model on the IMDB dataset, 50\% bottom corruption yields an F1 of 76.91, whereas 50\% top corruption results in 80.92. This observation aligns with the understanding that lower layers in BERT capture more fundamental linguistic features, and their corruption can have a more significant impact on the model's performance compared to corrupting the higher layers responsible for task-specific representations.
\begin{table*}[!h]
\centering
\caption{Fine-tuned BERT performance under different corruption settings across various datasets with the \textbf{average token}. Bold values represent higher values between bottom and top. Red values indicate outliers. Reported F1 score in percentage (\%)}
\resizebox{\linewidth}{!}
{
\begin{tabular}{cc|cc|cc|cc|cc|cc|cc}
\toprule
\multirow{2}{*}{Model} & \multirow{2}{*}{Corruption} & \multicolumn{2}{c}{SST-2} & \multicolumn{2}{c}{IMDB} & \multicolumn{2}{c}{AG News} & \multicolumn{2}{c}{Emotion} & \multicolumn{2}{c}{DBPedia} & \multicolumn{2}{c}{TFNT} \\ 

 &  & Bottom & \multicolumn{1}{c}{Top} & Bottom & \multicolumn{1}{c}{Top} & Bottom & \multicolumn{1}{c}{Top} & Bottom & \multicolumn{1}{c}{Top} & Bottom & \multicolumn{1}{c}{Top} & Bottom & \multicolumn{1}{c}{Top} \\  
 
\midrule
 
\multirow{5}{*}{Base} & 0\% & \multicolumn{2}{c|}{91.67} & \multicolumn{2}{c|}{82.31} & \multicolumn{2}{c|}{94.08} & \multicolumn{2}{c|}{93.20} & \multicolumn{2}{c|}{98.42} & \multicolumn{2}{c}{87.22} \\ 

 & 25\% (1-3) & 89.54 & \textbf{91.44} & 81.38 & \textbf{81.96} & 93.16 & \textbf{93.93} & 92.89 & \textbf{93.07} & 97.85 & \textbf{98.43} & 82.14 & \textbf{87.49} \\
 & 50\% (1-6) & 83.93 & \textbf{87.03} & 77.29 & \textbf{80.76} & 93.02 & \textbf{93.93} & 92.28 & \textbf{92.90} & 97.34 & \textbf{98.43} & 84.07 & \textbf{88.18} \\
 & 75\% (1-9) & \textbf{83.24} & 83.15 & 76.66 & \textbf{79.07} & 91.95 & \textbf{92.45} & 89.75 & \textbf{92.76} & 97.69 & \textbf{98.49} & 84.71 & \textbf{85.27} \\
 
 & 100\% (1-12) & \multicolumn{2}{c|}{77.07} & \multicolumn{2}{c|}{76.29} & \multicolumn{2}{c|}{90.71} & \multicolumn{2}{c|}{87.71} & \multicolumn{2}{c|}{94.56} & \multicolumn{2}{c}{76.53} \\ [0.5mm]
 
 \hhline{~~------------}
\rule{0pt}{4.3mm}

\multirow{5}{*}{Large} & 0\% & \multicolumn{2}{c|}{93.33} & \multicolumn{2}{c|}{82.31} & \multicolumn{2}{c|}{93.91} & \multicolumn{2}{c|}{93.48} & \multicolumn{2}{c|}{98.86} & \multicolumn{2}{c}{89.55} \\ 

 & 25\% (1-6) & 90.01 & \textbf{92.30} & 81.73 & \textbf{81.96} & 93.51 & \textbf{93.81} & 92.72 & \textbf{93.27} & 98.01 & \textbf{98.57} & 85.03 & \textbf{88.96} \\
 & 50\% (1-12) & \textcolor{red}{44.78} & \textbf{87.07} & 77.29 & \textbf{80.76} & 91.72 & \textbf{93.59} & 90.18 & \textbf{92.64} & 96.77 & \textbf{98.43} & 81.80 & \textbf{88.60} \\
 & 75\% (1-18) & 82.49 & \textbf{84.89} & 76.66 & \textbf{79.07} & 92.22 & \textbf{93.30} & 89.92 & \textbf{92.07} & 97.55 & \textbf{98.35} & 83.96 & \textbf{88.37} \\
 
 & 100\% (1-24) & \multicolumn{2}{c|}{78.66} & \multicolumn{2}{c|}{76.29} & \multicolumn{2}{c|}{90.47} & \multicolumn{2}{c|}{87.28} & \multicolumn{2}{c|}{93.57} & \multicolumn{2}{c}{79.36} \\ [0.5mm]
 
\hhline{~~------------}
\rule{0pt}{4.3mm}
 
\multirow{3}{*}{Distil} & 0\% & \multicolumn{2}{c|}{89.29} & \multicolumn{2}{c|}{81.67} & \multicolumn{2}{c|}{93.92} & \multicolumn{2}{c|}{93.25} & \multicolumn{2}{c|}{98.65} & \multicolumn{2}{c}{86.18} \\ 

 & 50\% (1-3) & 81.60 & \textbf{84.02} & 77.22 & \textbf{79.78} & 91.85 & \textbf{93.51} & 92.10 & \textbf{93.14} & 97.39 & \textbf{98.21} & 81.46 & \textbf{85.67} \\
 
 & 100\% (1-6) & \multicolumn{2}{c|}{77.48} & \multicolumn{2}{c|}{76.20} & \multicolumn{2}{c|}{90.50} & \multicolumn{2}{c|}{87.93} & \multicolumn{2}{c|}{95.47} & \multicolumn{2}{c}{75.64} \\
 
 \bottomrule
\end{tabular}

}

\label{tab:corruption_avg}
\end{table*}

\subsection{Further Exploration}

\paragraph{Average token instead of CLS token.}
We replace the [CLS] token in the BERT output mentioned in Section \ref{sec:bg} with the average of all non-[CLS] tokens output by BERT as the input of the classification matrix, and obtain the results in Table \ref{tab:corruption_avg}. These results further corroborate the three main conclusions drawn from the previous experiments. However, it is noteworthy that the results obtained using the average token strategy exhibit some fluctuations compared to the previous experiments using the [CLS] token. On the DBPedia dataset, we can observe some interesting anomalies in the base model. For 75\% top corruption achieves a score of 98.49\%, which is higher than both 25\% and 50\% top corruption. This deviates from our previous finding that higher degrees of corruption lead to worse performance. Despite these anomalies, the overall trend across multiple datasets and models still supports the general conclusions drawn earlier.  

These fluctuations suggest that the average token strategy may not be as robust as using the [CLS] token for representation, which is specifically designed for classification tasks. The [CLS] token captures task-specific information more effectively, leading to more consistent results across different corruption settings.

\begin{table}[!t]
\centering
\caption{Fine-tuned BERT linear probe performance under different corruption settings across various datasets. Bold values represent higher values between bottom and top. Red values indicate outliers. Reported F1 score in percentage (\%)}
\resizebox{\linewidth}{!}{
\begin{tabular}{cc|cc|cc}
\toprule
\multirow{2}{*}{Model} & \multirow{2}{*}{Corruption} & \multicolumn{2}{c}{AG News} & \multicolumn{2}{c}{DBPedia} \\ 
 &  & Bottom & \multicolumn{1}{c}{Top} & Bottom & \multicolumn{1}{c}{Top} \\ 
\midrule
\multirow{5}{*}{Base} & 0\% & \multicolumn{2}{c|}{81.41} & \multicolumn{2}{c}{77.33} \\ 
 & 25\% (1-3) & \textbf{79.73} & 78.41 & 76.15 & \textbf{83.27} \\
 & 50\% (1-6) & 67.53 & \textbf{80.58} & 59.85 & \textbf{80.61} \\
 & 75\% (1-9) & 69.93 & \textbf{82.18} & 66.49 & \textbf{82.54} \\
 & 100\% (1-12) & \multicolumn{2}{c|}{50.68} & \multicolumn{2}{c}{53.28} \\ [0.5mm]

 \hhline{~~----}
\rule{0pt}{4.3mm}

\multirow{5}{*}{Large} & 0\% & \multicolumn{2}{c|}{65.65} & \multicolumn{2}{c}{71.31} \\ 
 & 25\% (1-6) & 64.38 & \textbf{71.78} & 53.29 & \textbf{83.07} \\
 & 50\% (1-12) & \textcolor{red}{39.20} & \textbf{77.21} & \textcolor{red}{16.81} & \textbf{62.44} \\
 & 75\% (1-18) & 50.93 & \textbf{83.27} & \textcolor{red}{32.94} & \textbf{83.88} \\
 & 100\% (1-24) & \multicolumn{2}{c|}{51.19} & \multicolumn{2}{c}{57.74} \\ [0.5mm]
 
 \hhline{~~----}
\rule{0pt}{4.3mm}

\multirow{3}{*}{Distil} & 0\% & \multicolumn{2}{c|}{89.29} & \multicolumn{2}{c}{98.95} \\ 
 & 50\% (1-3) &82.98 & \textbf{88.11}& 91.94 & \textbf{97.19} \\
 & 100\% (1-6) & \multicolumn{2}{c|}{54.60} & \multicolumn{2}{c}{71.97} \\
\bottomrule
\end{tabular}
}

\label{tab:corruption_linear}
\end{table}

\begin{table}[!t]
\centering
\caption{Fine-tuned BERT performance under different corruption settings across non-classified datasets. Bold values represent higher values between bottom and top. Red values indicate outliers. Reported F1 score in percentage (\%)}
\resizebox{\linewidth}{!}{
\begin{tabular}{cc|cc|cc}
\toprule
\multirow{2}{*}{Model} & \multirow{2}{*}{Corruption} & \multicolumn{2}{c}{MRPC} & \multicolumn{2}{c}{RTE} \\ 
 &  & Bottom & \multicolumn{1}{c}{Top} & Bottom & \multicolumn{1}{c}{Top} \\ 
\midrule
\multirow{5}{*}{Base} & 0\% & \multicolumn{2}{c|}{76.38} & \multicolumn{2}{c}{55.84} \\ 
 & 25\% (1-3) & 64.73 & \textbf{73.58} & 54.22 & \textbf{60.85} \\
 & 50\% (1-6) & 65.00 & \textbf{73.68} & 53.71 & \textbf{57.78} \\
 & 75\% (1-9) & 63.10 & \textbf{67.15} & 53.34 & \textbf{57.20} \\
 & 100\% (1-12) & \multicolumn{2}{c|}{60.74} & \multicolumn{2}{c}{47.36} \\ [0.5mm]
 \hhline{~~----}
\rule{0pt}{4.3mm}
\multirow{5}{*}{Large} & 0\% & \multicolumn{2}{c|}{77.92} & \multicolumn{2}{c}{60.26} \\ 
 & 25\% (1-6) & 61.60 & \textbf{79.60} & 45.39 & \textbf{62.62} \\
 & 50\% (1-12) & \textcolor{red}{44.18} & \textbf{73.63} & \textcolor{red}{30.05} & \textbf{52.48} \\
 & 75\% (1-18) & 63.27 & \textbf{65.40} & \textcolor{red}{30.05} & 53.81 \\
 & 100\% (1-24) & \multicolumn{2}{c|}{63.27} & \multicolumn{2}{c}{\textcolor{red}{30.05}} \\ [0.5mm]

\hhline{~~----}
\rule{0pt}{4.3mm}
\multirow{3}{*}{Distil} & 0\% & \multicolumn{2}{c|}{79.03} & \multicolumn{2}{c}{56.12} \\ 
 & 50\% (1-3) & 64.49 & \textbf{71.00} & 57.82 & \textbf{60.66} \\
 & 100\% (1-6) & \multicolumn{2}{c|}{63.46} & \multicolumn{2}{c}{52.38} \\
\bottomrule
\end{tabular}
}

\label{tab:corruption_nonclassified}
\end{table}

\paragraph{Linear Probe.}
We have fine-tuned all BERT model parameters to assess the impact of parameter corruption on performance. Alternatively, we can use a linear probing approach by freezing the pre-trained BERT parameters and only fine-tuning the classification head. This method allows us to investigate how well corrupted BERT models transfer learned representations to downstream tasks without further adapting the main model parameters. The results presented in Table \ref{tab:corruption_linear} appear highly unstable and inconsistent, making it challenging to observe clear trends identified when the entire BERT is fine-tuned. One potential reason for this instability is that the optimization process for the classification head may struggle to converge, leading to erratic behavior and performance fluctuations. These irregularities and inconsistencies suggest that the linear probing approach may not be suited for accurately assessing the impact of parameter corruption on the model's performance. The lack of adaptation in the main BERT body could exacerbate the effects of corruption, leading to unstable and unpredictable results.

\paragraph{Non-classification Tasks.}
We also conduct some exploration of non-classified datasets. Specifically, we use the same method to test the RTE and MRPC datasets for text entailment tasks and similarity judgment tasks. The results are shown in Table \ref{tab:corruption_nonclassified}. While the overall trends observed in the previous experiments on classification tasks are also present in the results for non-classification tasks, they are much more volatile and inconsistent. For instance, on the MRPC dataset, the BERT-Base model's performance fluctuates significantly, with 25\% bottom corruption (54.22\%) outperforming the undamaged model (55.84\%), and 75\% bottom corruption (53.34\%) scoring higher than 50\% (53.71\%).

These irregularities and inconsistencies suggest that the trends observed in the classification tasks may not be as pronounced or robust in non-classification tasks. The nature of the tasks, the specific datasets, and the way the models encode and process the data for these tasks could all contribute to the increased volatility in performance under different corruption settings. It reminds us that the behavior of language models can be highly task-dependent, and the impact of parameter corruption may manifest differently across different types of tasks and datasets.

\subsection{Feature Similarity}
\begin{figure}[!t] 
    \centering
    \includegraphics[width=0.46\textwidth]{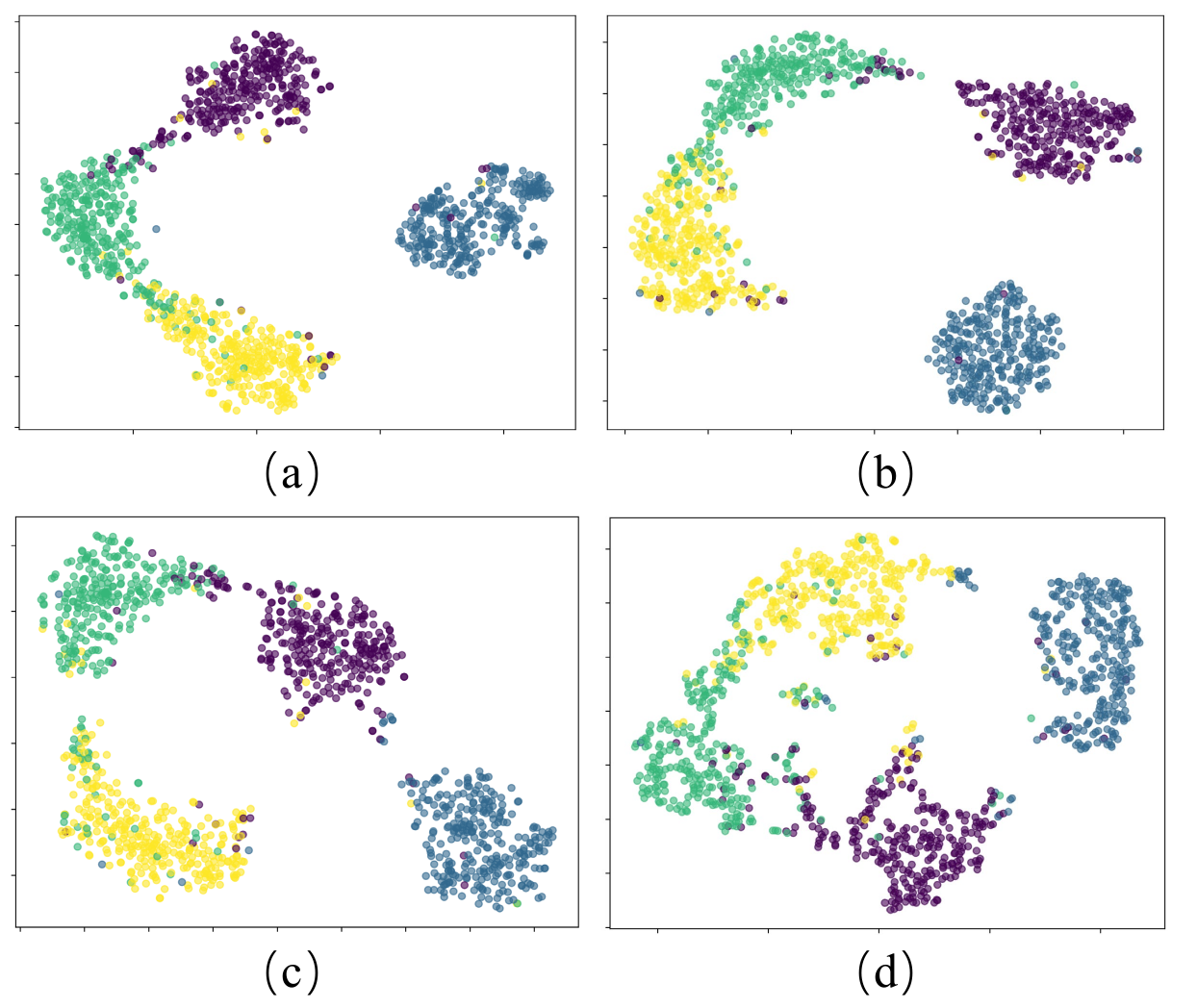} 
    \caption{Feature clustering visualizations of BERT-Base output with different corruption settings on AG News. (a) 0\% corruption (b) 25\% top corruption (c) 25\% bottom corruption (d) 100\% corruption}
    \label{fig:feature_vis}
    \vspace{-20pt}
\end{figure}
We employ the t-SNE \cite{van2008visualizing} algorithm to project [CLS] tokens gained from the fine-tuned BERT models with various corruption patterns to 2D space. t-SNE visualization reveals clusters in features, which help in understanding the classification performance and observing intrinsic feature patterns and relationships.

(a) 0\% corruption: In the undamaged BERT model, different classes form well-separated groups in 2D space, indicating the effective semantic distinction between categories.
(b) 25\% top corruption: Cluster structure is still visible, but some samples drift closer to other clusters, suggesting top layer corruption interferes with BERT's semantic representation.
(c) 25\% bottom corruption: Cluster boundaries become more blurred, with numerous samples overlapping with other categories' regions. This observation aligns with our quantitative findings, suggesting that bottom-layer corruption has a more significant impact on BERT's representational capabilities.
(d) 100\% corruption: In this scenario, the visualization suggests that severe parameter corruption has made the task of separating and clustering different semantic categories considerably more difficult for the model. However, some level of semantic differentiation may still be present, albeit obscured by the high degree of feature mixing and overlap caused by the extensive parameter corruption.

Overall, these visualizations corroborate our previous findings. As the degree of corruption increases, we observe a gradual transition from well-separated clusters to a completely mixed distribution, reflecting the degradation of the model's semantic understanding capabilities.
\section{Conclusion}
We study the robustness of BERT models to parameter corruption during fine-tuning for sentence classification tasks. By strategically corrupting model parameters, we find that corrupted models cannot fully recover original performance through fine-tuning, with higher corruption levels causing more severe performance degradation. Bottom-layer corruption has a greater impact than top-layer corruption. Our findings shed light on the interplay between pre-trained language models, fine-tuning, and parameter corruption, highlighting their sensitivity to corruption and limitations in recovering performance. These insights contribute to a deeper understanding of the robustness of language models and the precise effect of parameter corruption.
\section*{Impact Statements}
This paper presents work whose goal is to advance the field of Machine Learning. There are many potential societal consequences of our work, none of which we feel must be specifically highlighted here.

\nocite{langley00}

\bibliography{example_paper}
\bibliographystyle{icml2025}

\newpage
\appendix
\onecolumn

\section{More Analysis: Outlier Analysis.}
From our results in Table \ref{tab:corruption_cls}, intriguing outliers emerge, such as a low accuracy of 33.97\% for BERT-Large on IMDB with 50\% bottom corruption, and 16.98\% accuracy on Emotion with 25\% and 50\% bottom corruption. These anomalies, marked by convergence failures and low metrics, may result from the sensitivity of BERT-Large's middle layers, the complexity of its larger parameter set, and specific characteristics of the IMDB and Emotion datasets. These factors highlight the complex interplay between datasets, architectures, and corruption patterns, sometimes leading to unexpected outcomes.

\section{License of Scientific Artifacts}
We discuss the license of the used dataset as follows:

\paragraph{SST-2}
It uses a CC-zero License. Anyone can copy, modify, distribute, and perform the work, even for commercial purposes, all without asking permission. \href{https://creativecommons.org/publicdomain/zero/1.0/}{CC-zero Link}. We have cited the original paper.

\paragraph{IMDB}
There is no specific license. Anyone using the dataset should cite the \cite{maas-EtAl:2011:ACL-HLT2011}

\paragraph{AGNews}
AGNews has \href{http://groups.di.unipi.it/~gulli/AG_corpus_of_news_articles.html}{Custom (non-commercial)} License. The dataset is provided by the academic community for research purposes in data mining (clustering, classification, etc), information retrieval (ranking, search, etc), XML, data compression, data streaming, and any other non-commercial activity. Everyone is encouraged to download this corpus for any non-commercial use. 

\paragraph{Emotion}
The \href{https://huggingface.co/datasets/dair-ai/emotion}{Hugging Face page} shows this dataset should be used for educational and research purposes only.

\paragraph{DBPedia}
The DBPedia ontology classification dataset is licensed under the terms of the Creative Commons Attribution-ShareAlike License and the GNU Free Documentation License.

\paragraph{Twitter Financial News Topic (TFNT)}
The Twitter Financial Dataset (topic) version 1.0.0 is released under the MIT License.

\paragraph{MRPC}
There is no specific license in this \href{https://www.microsoft.com/en-us/download/details.aspx?id=52398}{page}. Anyone using the dataset should cite the \cite{Ando2005}.

\paragraph{RTE}
The Recognizing Textual Entailment (RTE) datasets come from a series of annual textual entailment challenges. The authors of the GLUE benchmark combined the data from RTE1 \cite{dagan2006pascal}, RTE2 \cite{bar2006second}, RTE3 \cite{giampiccolo2007third} and RTE5 \cite{bentivogli2009fifth}. Examples are constructed based on news and Wikipedia text. Anyone using the dataset should cite above and GLUE \cite{wang2019glue}. There is no specific license in \href{https://tac.nist.gov//data/}{tac page} and \href{https://aclweb.org/aclwiki/Recognizing_Textual_Entailment}{acl page}.

\section{Limitation}
Here are a few limitations of this work:
\begin{itemize}
    \item Corruption Method: Our study employs a specific method of parameter corruption by random initialization. Other types of corruption, such as noise injection or targeted parameter manipulation, can potentially yield different insights. 
    \item Model Architectures: The research focuses on the BERT family of language models. The impact of parameter corruption may vary across different model architectures, such as recurrent neural networks, convolutional models, or other Transformer-based models. 
    \item Fine-tuning Strategies: The study explores the standard fine-tuning approach, where all model parameters are updated during training. Alternative fine-tuning strategies, such as layer-wise adaptation or low-rank adaptation, could potentially mitigate or exacerbate the effects of parameter corruption.
\end{itemize}

\end{document}